# Automatic Brain Tumor Detection and Segmentation Using U-Net Based Fully Convolutional Networks

Hao Dong [†1], Guang Yang [†2,3], Fangde Liu [†1], Yuanhan Mo [1], Yike Guo [‡1]

[1] Data Science Institute, Imperial College London, SW7 2AZ, London, UK
[2] Neurosciences Research Centre, Molecular and Clinical Sciences Institute,
St. George's, University of London, London SW17 0RE, UK
[3] National Heart and Lung Institute, Imperial College London, SW7 2AZ, London, UK
[†] Co-first authors contributed equally to this study.
[‡] Corresponding author.
{hao.dong11, g.yang, fangde.liu, y.mo16, y.guo}@imperial.ac.uk

**Abstract.** A major challenge in brain tumor treatment planning and quantitative evaluation is determination of the tumor extent. The noninvasive magnetic resonance imaging (MRI) technique has emerged as a front-line diagnostic tool for brain tumors without ionizing radiation. Manual segmentation of brain tumor extent from 3D MRI volumes is a very time-consuming task and the performance is highly relied on operator's experience. In this context, a reliable fully automatic segmentation method for the brain tumor segmentation is necessary for an efficient measurement of the tumor extent. In this study, we propose a fully automatic method for brain tumor segmentation, which is developed using U-Net based deep convolutional networks. Our method was evaluated on Multimodal Brain Tumor Image Segmentation (BRATS 2015) datasets, which contain 220 high-grade brain tumor and 54 low-grade tumor cases. Cross-validation has shown that our method can obtain promising segmentation efficiently.

## 1 Introduction

Primary malignant brain tumors are among the most dreadful types of cancer, not only because of the dismal prognosis, but also due to the direct consequences on decreased cognitive function and poor quality of life. The most frequent primary brain tumors in adults are primary central nervous system lymphomas and gliomas, which the latter account for almost 80% of malignant cases [1]. The term glioma encompasses many subtypes of the primary brain tumor, which range from slower-growing 'low-grade' tumors to heterogeneous, highly infiltrative malignant tumors. Despite significant advances in imaging, radiotherapy, chemotherapy and surgical procedure, certain cases of malignant brain tumors, e.g., high-grade glioblastoma and metastasis, are still considered untreatable with a 2.5-year cumulative relative survival rate of 8% and 2% at 10 years [2]. Moreover, there are variable prognosis results for patients with low-grade gliomas (LGG) with an overall 10-year survival rate about 57% [3].

Previous studies have demonstrated that the magnetic resonance imaging (MRI) characteristics of newly identified brain tumors can be used to indicate the likely diagnosis and treatment strategy [4–6]. In addition, multimodal MRI protocols are normally used to evaluate brain tumor cellularity, vascularity, and blood-brain barrier (BBB) integrity. This is because different image contrasts produced by multimodal MRI protocols can provide crucial complementary information. Typical brain tumor MRI protocols, which are used routinely, include $T_1$-weighted, $T_2$-weighted (including Fluid-Attenuated Inversion Recovery, i.e., FLAIR), and gadolinium enhanced $T_1$-weighted imaging sequences. These structural MRI images yield a valuable diagnosis in the majority of cases [7].

Image segmentation is a critical step for the MRI images to be used in brain tumor studies: (1) the segmented brain tumor extent can eliminate confounding structures from other brain tissues and therefore provide a more accurate classification for the sub-types of brain tumors and inform the subsequent diagnosis; (2) the accurate delineation is crucial in radiotherapy or surgical planning, from which not only brain tumor extend has been outlined and surrounding healthy tissues has been excluded carefully in order to avoid injury to the sites of language, motor, and sensory function during the therapy; and (3) segmentation of longitudinal MRI scans can efficiently monitor brain tumor recurrence, growth or shrinkage. In current clinical practice, the segmentation is still relied on manual delineation by human operators. The manual segmentation is a very labor-intensive task, which normally involves slice-by-slice procedures, and the results are greatly dependent on operators' experience and their subjective decision making. Moreover, reproducible results are difficult to achieve even by the same operator. For a multimodal, multi-institutional and longitudinal clinical trial, a fully automatic, objective and reproducible segmentation method is highly in demand.

Despite recent developing in semi-automatic and fully automatic algorithms for brain tumor segmentation, there are still several opening challenges for this task mainly due to the high variation of brain tumors in size, shape, regularity, location and their heterogeneous appearance (e.g., contrast uptake, image uniformity and texture) [6, 8]. Other potential issues that may complicate the brain tumor segmentation include: (1) the BBB normally remains intact in LGG cases and the tumor regions are usually not contrast enhanced; therefore, the boundaries of LGG can be invisible or blurry despite FLAIR sequence may provide differentiation between normal brain and brain tumor or edema to delineate the full extent of the lesion; (2) in contrast, for high-grade gliomas (HGG) cases, the contrast agent, e.g., gadolinium, leaks across the disrupted BBB and enters extracellular space of the brain tumor causing hyper-intensity on $T_1$-weighted images. Therefore, the necrosis and active tumor regions can be easily delineated. However, HGG usually exhibit unclear and irregular boundaries that might also involve discontinuities due to aggressive tumor infiltration. This can cause problems and result in poor tumor segmentation; (3) varies tumor sub-regions and tumor types can only be visible by considering multimodal MRI data. However, the co-registration across multiple MRI sequences can be difficult especially when these sequences are acquired in different spatial resolutions; and (4) typical clinical MRI images are normally acquired with higher in-plane resolution and much lower inter-slice resolution in order to balance between adequate image slices to cover the whole tumor volume with good quality cross-sectional views and the restricted scanning time. This can cause inadequate signal to noise ratio and asymmetrical partial volume effects may also affect the final segmentation accuracy.

Previous studies on brain tumor segmentation can be roughly categorized into un-supervised learning based [9–12] and supervised learning based [13–18] methods. A more detailed topical review on various brain tumor segmentation methods can be found elsewhere, e.g., in [6]. In addition, a dedicated annual workshop and challenge, namely Multimodal Brain Tumor Image Segmentation (BRATS), is held to benchmark different algorithms that developed for the brain tumor segmentation [19]. Here we only reviewed some most recent and closely relevant studies for this topic.

Unsupervised learning based clustering has been successfully used for brain tumor segmentation by grouping data based on certain similarity criteria. Hsieh et al. [20] combined fuzzy clustering with region-growing for brain tumor cases scanned by $T_1$-weighted and $T_2$-weighted sequences and achieved a segmentation accuracy of 73%. In [9], a multi-stage fuzzy c-means framework was proposed to segment brain tumors scanned by multimodal MRI and obtained promising results, but the proposed framework was tested on a very limited number of datasets. Recently, a study [11] has been carried out to evaluate different clustering algorithms for glioblastoma segmentation, and results showed that Gaussian hidden Markov random field outperformed k-means, fuzzy k-means and Gaussian mixture model for this task. However, the best performing algorithm described in this study still only achieved 77% accuracy.

On the other hand, supervised learning based methods require training data-and-label pairs to learn a classification model, based on which new instances can be classified and then segmented. Wu et al. [13] employed superpixel features in a conditional random fields framework to segment brain tumors, but the results varied significantly among different patient cases and especially underperformed in LGG images. A study was proposed in which extremely randomized forest was used for classifying both appearance and context based features and 83% Dice score was achieved [14]. More recently, Soltaninejad et al. [16] combined extremely randomized trees classification with superpixel based over-segmentation for a single FLAIR sequence based MRI scan that obtained 88% overall Dice score of the complete tumor segmentation for both LGG and HGG tumor cases. Nevertheless, the tuning of superpixel size and compactness could be tricky and influence the final delineation.

Recently, supervised deep convolutional neural networks (CNN) have attracted lots of interests. Compared to conventional supervised machine learning methods, these deep learning based methods are not dependent on hand-crafted features, but automatically learn a hierarchy of increasingly complex features directly from data [21]. Currently, using BRATS datasets and their benchmarking system, deep learning based methods have been ranked on top of the contest [21–23]. This can be attributed to the fact that deep CNN is constructed by stacking several convolutional layers, which involve convolving a signal or an image with kernels to form a hierarchy of features that are more robust and adaptive for the discriminative models. Despite recent advances in these deep learning based methods, there are still several challenges: (1) essentially tumor segmentation is an abnormal detection problem, it is more challenging than other pattern recognition based tasks; (2) while most methods provided satisfied segmentation for HGG cases, in general the performance of the LGG segmentation is still poor; (3) compared to complete tumor segmentation, the delineation of core tumor regions and enhanced infiltrative regions is still underperformed; (4) a more computing-efficient and memory-efficient development is still in demand be-cause existing CNN based methods require considerable amount of computing resources.

In this study, we developed a novel 2D fully convoluted segmentation network that is based on the U-Net architecture [24]. In order to boost the segmentation accuracy, a comprehensive data augmentation technique has been used in this work. In addition, we applied a 'Soft' Dice based loss function introduced in [25]. The Soft Dice based loss function has a unique advantage that is adaptive to unbalanced samples, which is very important for brain tumor segmentation because some sub-tumoral regions may only count for a small portion of the whole tumoral volume. The proposed method has been validated using datasets acquired for both LGG and HGG patients. Compared with manual delineated ground truth, our fully automatic method has obtained promising results. Also compared to other state-of-the-art methods, we have achieved comparable results for delineating the complete tumor regions, and superior segmentation for the core tumor regions.

## 2    Method

### 2.1    Brain MRI Data Acquisitions and Patients

The proposed method was tested and evaluated on the BRATS 2015 datasets [19], which contain 220 high-grade glioma (HGG) and 54 low-grade glioma (LGG) patient scans. Multimodal MRI data is available for every patient in the BRATS 2015 datasets and four MRI scanning sequences were performed for each patient using $T_1$-weighted (T1), $T_1$-weighted imaging with gadolinium enhancing contrast (T1c), $T_2$-weighted (T2) and FLAIR. For each patient, the T1, T2 and FLAIR images were co-registered into the T1c data, which had the finest spatial resolution, and then resampled and interpolated into $1\times1\times1mm^3$ with an image size of 240×240×155. We have applied data normalization for each sequence of the multimodal MRI by subtracting the mean of each sequence and dividing by its standard deviation.

In addition, manual segmentations with four intra-tumoral classes (labels) are available for each case: necrosis (1), edema (2), non-enhancing (3), and enhancing tumor (4). The manual segmentations have been used as the ground truth in both segmentation model training and final segmentation performance evaluation. In previous studies, multimodal data were stacked like the multichannel RGB images [21–23]. In this study, we used FLAIR images to segment the complete tumor regions and tumor regions except edema that has been proved to be effective [16]. Additionally, T1c data were used to delineate the enhancing tumor. In so doing, our framework is not only more efficient, but also requires less clinical inputs because frequently multimodal MRI data are not available due to patient symptoms and limited acquisition time.

### 2.2    Data Augmentation

The purpose of data augmentation is to improve the network performance by intentionally producing more training data from the original one. In this study, we applied a set of data augmentation methods summarized in **Table 1**. Simple transformation such as flipping, rotation, shift and zoom can result in displacement fields to images but will not create training samples with very different shapes. Shear operation can slightly distort the global shape of tumor in the horizontal direction, but is still not powerful to gain sufficient variable training data, as tumors have no definite shape. To cope with this problem, we further applied elastic distortion [26] that can generate more training data with arbitrary but reasonable shapes.

**Table 1.** Summary of the applied data augmentation methods ($\gamma$ controls the brightness of the outputs; $\alpha$ and $\sigma$ control the degree of the elastic distortion).

| Methods | Range |
| --- | --- |
| Flip horizontally | 50% probability |
| Flip vertically | 50% probability |
| Rotation | $\pm 20^\circ$ |
| Shift | 10% on both horizontal and vertical direction |
| Shear | 20% on horizontal direction |
| Zoom | $\pm 10\%$ |
| Brightness | $\gamma=0.8\sim1.2$ |
| Elastic distortion | $\alpha=720$, $\sigma=24$, |

### 2.3 U-Net Based Deep Convolutional Networks

Biomedical images usually contain detailed patterns of the imaged object (e.g., brain tumor), and the edge of the object is variable. To cope with the segmentation for the objects with detailed patterns, Long et al. [27] proposed to use the skip-architecture that combined the high-level representation from deep decoding layers with the appearance representation from shallow encoding layers to produce detailed segmentation. This method has demonstrated promising results on natural images [27] and is also applicable to biomedical images [28]. Ronneberger et al. [24] introduced the U-Net, which employed the skip-architecture, to solve the cell tracking problem.

Our network architecture, which is based on the U-Net, consists of a down-sampling (encoding) path and an up-sampling (decoding) path as shown in **Fig. 1**. The down-sampling path has 5 convolutional blocks. Every block has two convolutional layers with a filter size of 3×3, stride of 1 in both directions and rectifier activation, which increase the number of feature maps from 1 to 1024. For the down-sampling, max pooling with stride 2×2 is applied to the end of every blocks except the last block, so the size of feature maps decrease from 240×240 to 15×15. In the up-sampling path, every block starts with a deconvolutional layer with filter size of 3×3 and stride of 2×2, which doubles the size of feature maps in both directions but decreases the number of feature maps by two, so the size of feature maps increases from 15×15 to 240×240. In every up-sampling block, two convolutional layers reduce the number of feature maps of concatenation of deconvolutional feature maps and the feature maps from encoding path. Different from the original U-Net architecture [24], we use zero padding to keep the output dimension for all the convolutional layers of both down-sampling and up-sampling path. Finally, a 1×1 convolutional layer is used to reduce the number of feature maps to two that refect the foreground and background segmentation respectively. No fully connected layer is invoked in the network. Other parameters of the network are tabulated in **Table 2**.

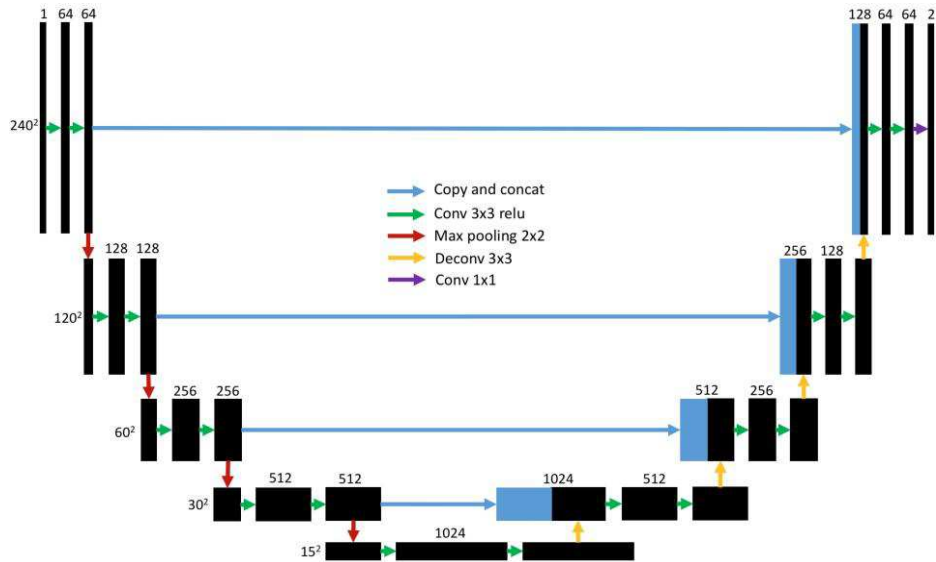

**Fig. 1.** Our developed U-Net architecture.

**Table 2.** Parameters setting for the developed U-Net.

| Parameters | Value |
| --- | --- |
| Number of convolutional blocks | [4, 5, 6] |
| Number of deconvolutional blocks | [4, 5, 6] |
| Regularization | L1, L2, Dropout |

### 2.4   Training and Optimization

During the training process, the Soft Dice metric described in [25] was used as the cost function of the network rather than the cross-entropy based or the quadratic cost function. Soft Dice can be considered as a differentiable form of the original Dice Similarity Coefficient (DSC) [25].

Training deep neural networks requires stochastic gradient-based optimization to minimize the cost function with respect to its parameters. We adopted the adaptive moment estimator (Adam) [29] to estimate the parameters. In general, Adam utilizes the first and second moments of gradients for updating and correcting moving average of the current gradients. The parameters of our Adam optimizer were set as: learning rate = 0.0001 and the maximum number of epochs = 100. All weights were initialized by normal distribution with mean of 0 and standard deviation of 0.01, and all biases were initialized as 0.

### 2.5 Experiments and Performance Evaluation

The evaluation has been done using a five-fold cross-validation method for the HGG and LGG data, respectively. For each patient, we have validated on three sub-tumoral regions as described by
   a. The *complete* tumor region (including all four intra-tumoral classes, labels 1, 2 3, and 4).
   b. The *core* tumor region (as above but excluded "edema" regions, labels 1, 3, and 4).
   c. The *enhancing* tumor region (only label 4).

For each tumoral region, the segmentations have been evaluated using the DSC, and the Sensitivity was also calculated. The DSC provides the overlap measurement between the manual delineated brain tumoral regions and the segmentation results of our fully automatic method that is

$$\text{DSC} = \frac{2\text{TP}}{\text{FP} + 2\text{TP} + \text{FN}}, \quad (1)$$

in which TP, FP and FN denote the true positive, false positive and false negative measurements, respectively.

In addition, Sensitivity is used to evaluate the number of TP and FN that is

$$\text{Sensitivity} = \frac{\text{TP}}{\text{TP} + \text{FN}}. \quad (2)$$

We reported the mean DSC results of the five-fold cross-validation and also showed boxplots of the corresponding sensitivities. In this study, the HGG and LGG cases are trained and cross-validated separately.

## 3  Results and Discussion

In this study, we proposed and developed U-Net based fully convolutional networks for solving the brain tumor segmentation problem. Essentially, tumor detection and segmentation belongs to the task of semantic segmentation. Compared to previous deep learning based studies on this topic, we employed a comprehensive data augmentation scheme that not only contains rigid or affine based deformation, but also includes brightness and elastic distortion based transformation, and this has then been coupled with the U-Net that incorporates the skip-architecture.

**Table 3** tabulates the DSC results of our cross-validated segmentation results for the HGG and LGG cases, respectively. In our current study, we only compared with three different deep learning based studies that published recently. All these three studies currently ranked on the top of the BRATS challenge. To the best of our knowledge, most published full papers were still focused on the BRATS 2013 datasets, which contains much less patient cases than the BRATS 2015 datasets. For example, in [21, 22] the model building has been done on the BRATS 2013 training datasets and then tested on the BRATS 2015 challenge datasets. Compared to the results on the BRATS 2015 challenge datasets [21, 22], the cross validation demonstrated that our method obtained superior results for the complete and core tumor segmentations. By using our method, the enhancing tumor segmentation for the LGG cases by using the T1c images only is not successful. This may be attributed to three reasons: (1) the BBB remains intact in most of these LGG cases and the tumor regions are rarely contrast enhanced; (2) the LGG cohort contains only 54 cases and the training datasets might be insufficient and (3) in these LGG cases, the borders between enhanced tumor and non-enhanced regions are more diffused and less visible that causes problems for both manual delineated ground truth and our fully automated segmentation model. Nevertheless, our method achieved 0.81 DSC for the enhancing tumor segmentation in the HGG cohort. Compared to Kamnitsas et al.'s work on the BRATS 2015 datasets [23], our method obtained comparable complete tumor segmentation results while achieving higher DSC for the core tumor delineation. **Fig. 2** displays the boxplots of the calculated Sensitivities and **Fig. 3** shows some exemplar qualitative overlaid segmentation results compared to the ground truth.

**Table 3.** Quantitative results of our proposed fully automatic brain tumor segmentation method compared to the results from other recently published deep learning based methods. Here we tabulated the Dice Similarity Coefficient (DSC) for HGG, LGG and combined cases, respectively. Grey background highlighted the experiments on the BRATS 2015 datasets. Bold numbers highlighted the results of the best performing algorithm.

| | | | DSC | | |
|---|---|---|---|---|---|
| **Method** | **Data** | **Grade** | **Complete** | **Core** | **Enhancing** |
| **Proposed** | Cross-Validation on BRATS 2015 Training Datasets | HGG | 0.88 | 0.87 | 0.81 |
| | | LGG | 0.84 | 0.85 | 0.00 |
| | | Combined | 0.86 | **0.86** | 0.65 |
| Pereira16 | BRATS 2013 Leaderboard | HGG | 0.88 | 0.76 | 0.73 |
| | | LGG | 0.65 | 0.53 | 0.00 |
| | | Combined | 0.84 | 0.72 | 0.62 |
| | BRATS 2013 Challenge | HGG | 0.88 | 0.83 | 0.77 |
| | BRATS 2015 Challenge | Combined | 0.79 | 0.65 | 0.75 |
| Havaei16 | BRATS 2013 Training | Combined | 0.88 | 0.79 | 0.73 |
| | BRATS 2013 Challenge | Combined | 0.88 | 0.79 | 0.73 |
| | BRATS 2013 Leaderboard | Combined | 0.84 | 0.71 | 0.57 |
| | BRATS 2015 Challenge | Combined | 0.79 | 0.58 | 0.69 |
| Kamnitsas17 | BRATS 2015 Training | Combined | **0.90** | 0.76 | **0.73** |
| | BRATS 2015 Challenge | Combined | 0.85 | 0.67 | 0.63 |

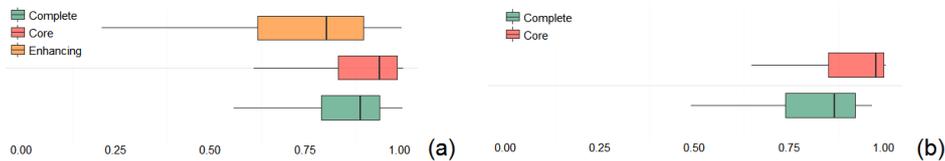

**Fig. 2.** Boxplots of the Sensitivity; (a) for the HGG cases and (b) for the LGG cases.

Our network has been implemented using the TensorFlow and the TensorLayer libraries. For the HGG cohort, each cross validation training session requires approximately 18 hours to finish on an NVIDIA Titan X (Pascal) graphics processing unit (GPU) with 12G memory, while the LGG cohort takes about ¼ of the training time of the HGG cohort. There is a trade-off in choosing between 2D and 3D models. Due to the memory limits of the GPU, a 2D U-Net model can process a full slice in one go while a 3D convolution system can only process a small patch cover a small portion of the 3D volume. Therefore, in this study we used a 2D based network.

For the prediction once we fixed the model, the computation time is approximately 2 to 3 seconds per case regardless a HGG or a LGG study. Compared to our computational time, previous studies were less computational efficient: ~30 seconds [23], 25 seconds to 3 minutes [22], and about 8 minutes [21] to predict one tumor study.

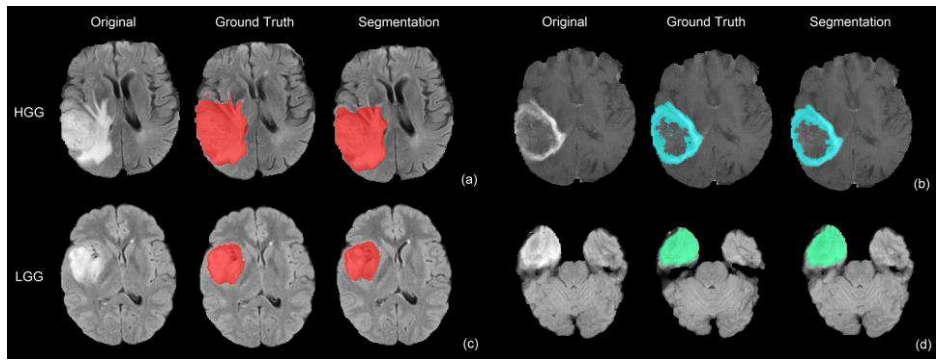

**Fig. 3.** Segmentations results for the exemplar HGG and LGG cases compared with manual delineated ground truth; (a) segmented complete tumor (red) of a HGG case overlaid on the FLAIR image; (b) segmented enhancing tumor (cyan) of a HGG case overlaid on the T1c image; (c) segmented complete tumor (red) of a LGG case overlaid on the FLAIR image; (a) segmented core tumor (green) of a LGG case overlaid on the FLAIR image.

There are still some limitations of the current work. First, our segmentation method has been evaluated using a cross-validation scheme, which can provide an unbiased predictor, but running our model on a separate and independent testing dataset may produce a more objective evaluation. Secondly, there are several parameters need to be carefully tuned in our network. Currently all the parameters were determined via empirical study. In particular, for the regularization, we did not find a significant performance improvement after applying L1, L2 or dropout to the network. This may be attributed to the fact that an effective image distortion has been applied during our model training, and it is difficult to overfit a network with large amount of training data. Moreover, our current framework is less successful in segmenting enhancing tumor regions of the LGG cohort. By stacking all the multimodal MRI channels and performing joint training with HGG datasets may solve the problem. Despite these limitations, the developed model has still demonstrated promising segmentation results with efficiency. We can certainly envisage its application and effectiveness on an independent testing dataset. In addition, validating our method on multi-institutional and longitudinal datasets, especially for clinical datasets with anisotropic resolutions, will be one of the future directions.

## 4     Conclusion

In this paper, we presented a fully automatic brain tumor detection and segmentation method using the U-Net based deep convolution networks. Based on the experiments on a well-established benchmarking (BRATS 2015) datasets, which contain both HGG and LGG patients, we have demonstrated that our method can provide both efficient and robust segmentation compared to the manual delineated ground truth. In addition, compared to other state-of-the-art methods, our U-Net based deep convolution networks can also achieve comparable results for the complete tumor regions, and superior results for the core tumor regions. In our current study, the validation has been carried out using a five-fold cross-validation scheme; however, we can envisage a straightforward application on an independent testing datasets and further applications for multi-institutional and longitudinal datasets. The proposed method makes it possible to generate a patient-specific brain tumor segmentation model without manual interference, and this potentially enables objective lesion assessment for clinical tasks such as diagnosis, treatment planning and patient monitoring.